\documentclass[11pt,a4paper]{article}
\usepackage{times,latexsym}
\usepackage{url}
\usepackage[T1]{fontenc}

\usepackage[acceptedWithA]{tacl2021v1}
\usepackage{algorithmic}
\usepackage{algorithm}
\usepackage{multirow}

\usepackage[pdftex]{graphicx}
\usepackage{caption}
\usepackage{subcaption}

\usepackage{amsmath}
\usepackage{amsthm}
\usepackage{amsfonts}
\theoremstyle{definition}
\newtheorem{definition}{Definition}[] 

\newcounter{subdefinition}[definition]
\renewcommand{\thesubdefinition}{\thedefinition.\arabic{subdefinition}}
\newenvironment{subdefinition}{
        \refstepcounter{subdefinition}
        \par\noindent
        \textbf{\upshape Definition \thesubdefinition}%
}{}

\usepackage{xspace,mfirstuc,tabulary}

\newcommand{\ex}[1]{{\sf #1}}

\newif\iftaclinstructions
\taclinstructionsfalse 
\iftaclinstructions
\renewcommand{\confidential}{}
\renewcommand{\anonsubtext}{(No author info supplied here, for consistency with
TACL-submission anonymization requirements)}
\newcommand{\instr}
\fi

\iftaclpubformat 

\else

\fi

\title{Self-supervised Topic Taxonomy Discovery in the Box Embedding Space}

\author{
  Yuyin Lu$^\diamondsuit$, Hegang Chen$^\diamondsuit$, Pengbo Mao$^\diamondsuit$, Yanghui Rao\Thanks{{The corresponding author.}}~ $^\diamondsuit$,\\
  \textbf{Haoran Xie$^\heartsuit$, Fu Lee Wang$^\clubsuit$,}
  \and
  \textbf{Qing Li$^\spadesuit$}
  \\
  $^\diamondsuit$School of
Computer Science and Engineering, Sun Yat-sen University, Guangzhou, China
  \\
  $^\heartsuit$School of Data Science, Lingnan University, Hong Kong SAR
  \\
  $^\clubsuit$School of Science and Technology, Hong Kong Metropolitan University, Hong Kong SAR
  \\
  $^\spadesuit$Department of Computing, the Hong Kong Polytechnic University, Hong Kong SAR
  \\
  \texttt{\{luyy37,chenhg25,maopb\}@mail2.sysu.edu.cn,}
  \\
  \texttt{raoyangh@mail.sysu.edu.cn,hrxie@ieee.org,}
  \\
  \texttt{pwang@hkmu.edu.hk,csqli@comp.polyu.edu.hk}
}

\date{}

\begin{document}
\maketitle
\begin{abstract}
  Topic taxonomy discovery aims at uncovering topics of different abstraction levels and constructing hierarchical relations between them. Unfortunately, most of prior work can hardly model semantic scopes of words and topics by holding the Euclidean embedding space assumption. What's worse, they infer asymmetric hierarchical relations by symmetric distances between topic embeddings. As a result, existing methods suffer from problems of low-quality topics at high abstraction levels and inaccurate hierarchical relations. To alleviate these problems, this paper develops a Box embedding-based Topic Model (BoxTM) that maps words and topics into the box embedding space, where the asymmetric metric is defined to properly infer hierarchical relations among topics. Additionally, our BoxTM explicitly infers upper-level topics based on correlation between specific topics through recursive clustering on topic boxes. Finally, extensive experiments validate high-quality of the topic taxonomy learned by BoxTM.
\end{abstract}

\iftaclpubformat
\section{Introduction}
\label{sec:introduction}

Taxonomy knowledge discovery, the process of extracting latent semantic hierarchies from text corpora, is a crucial while challenging research field. For text mining applications, it can serve as the foundation of complex question answering \cite{luo2018knowledge} and recommendation systems \cite{xie2022contrastive}. An important line of research focuses on learning word-level or entity-level taxonomies \cite{miller1995wordnet,jiang2022taxoenrich}, but such products may encounter problems of low coverage, high redundancy, and limited information \cite{zhang2018taxogen}. Since a topic can cover the semantics of a set of coherent words, some works propose to use topics as the basic taxonomic units. Taking the topic taxonomy of the arXiv website as an example, ``\textit{computer science}'' is an academic discipline highlighted by general keywords of ``\textit{information}'', ``\textit{computation}'', and ``\textit{automation}''. It involves various sub-fields such as ``\textit{computation and language}'' and ``\textit{computer vision}'', which have specific keywords of ``\textit{language}'' and ``\textit{image}'', respectively. With this topic taxonomy, users can readily retrieve papers of interest and explore related research fields.

Early methods for topic taxonomy discovery \cite{blei2003hierarchical,kim2012modeling,mimno2007mixtures} take a probabilistic perspective originated from LDA \cite{blei2003latent}. In these approaches, each topic is a distribution across words. A document is generated by sampling topics in different levels, and then sampling words from the selected topics iteratively. As a more flexible and efficient solution compared with probabilistic models, the Hierarchical Neural Topic Models (HNTMs) that adopt deep generative models and Neural Variational Inference (NVI) have been developed in recent years \cite{isonuma2020tree}. With remarkable developments of text representation learning \cite{pennington2014glove,devlin2019bert,vilnis2021geometric}, mining topic taxonomy in the high-quality embedding space has become a promising idea. Particularly, the latest HNTMs \cite{chen2021tree,duan2021sawtooth} extend the Embedded Topic Modeling (ETM) \cite{dieng2020topic} method to topic taxonomy discovery. With the assumption that topics and their keywords are close in the embedding space, these models utilize dot products between topic and word embeddings to infer topic-word distributions. 

In parallel, some other methods conduct recursive clustering on word embeddings to construct topic taxonomy directly \cite{zhang2018taxogen,grootendorst2022bertopic}. Such clustering-based methods often train the word embedding space on local contexts, which helps them capture accurate word semantics. Unfortunately, they have difficulty in exploiting global statistics of word occurrences, such as Bag-of-Words and TF-IDF representations. As a result, topics mined by these methods are highly coherent but may not be representative of the entire corpus. Due to this flaw of clustering-based methods, HNTMs persist as the prevailing paradigm for topic taxonomy discovery.

Despite the impressive performance of existing HNTMs, they suffer from the following problems. \textbf{(1) \textit{Suboptimal representations}:} Most of these methods are limited in modeling semantic scopes of words and topics at different abstraction levels using classic point embeddings \cite{pennington2014glove}. Instead, geometric embeddings such as hyperbolic and box embeddings are more effective representations for structured data, including knowledge graphs and taxonomies  \cite{bai2021modeling,abboud2020boxe}. Although HyperMiner \cite{shi2022hyperminer} attempts to uncover topic taxonomy within a geometric embedding space, it simply replaces point embeddings in traditional HNTMs with hyperbolic embeddings and lacks in-depth analysis. This makes HyperMiner suffer from the following problems. \textbf{(2) \textit{Topic collapse}:} prior models struggle to learn high-quality topics, especially at higher abstraction levels. In particular, their top-level topics often degenerate into clusters of meaningless common words \cite{wang2023hierarchical,wu2023effective}. \textbf{(3) \textit{Inaccurate hierarchy relations}:} many existing HNTMs rely on the symmetric distance metric (i.e., dot product) to infer the asymmetric hierarchy relations among topics. Such approximation results in an inaccurate hierarchical topic structure.

\begin{figure}[!t]
\centering
\includegraphics[width=\linewidth]{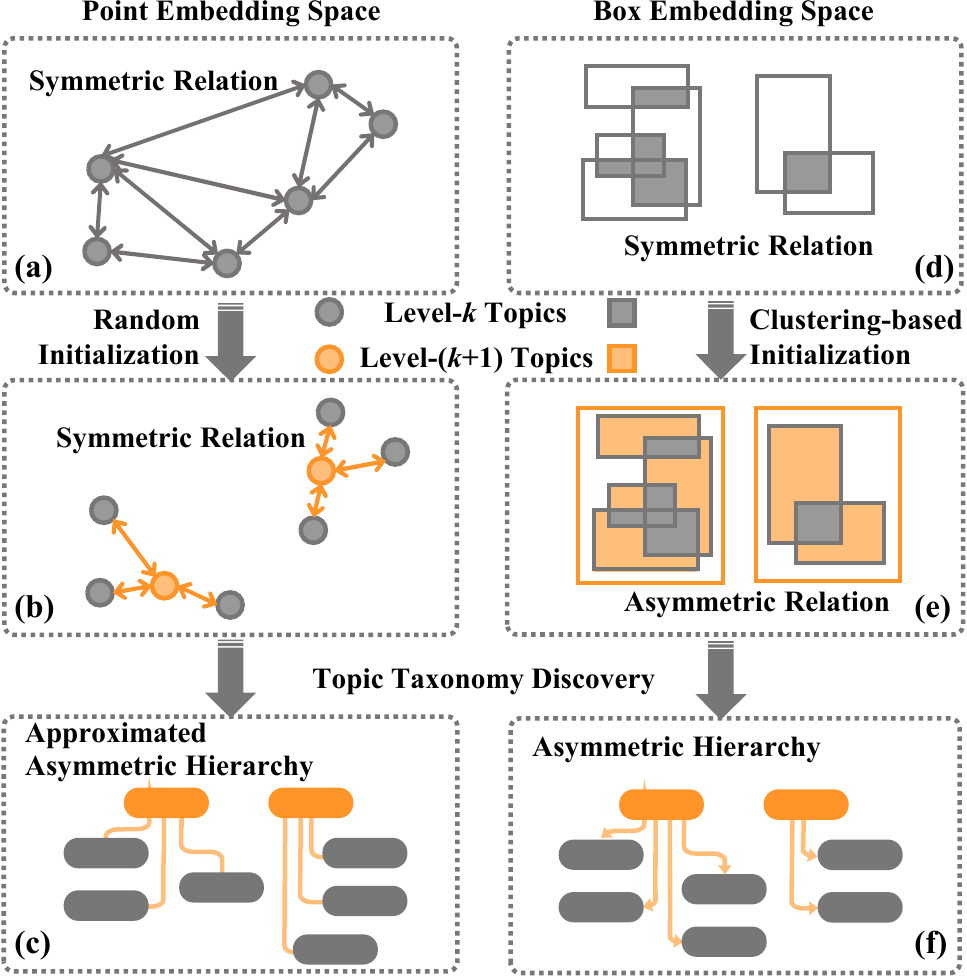}
\caption{The topic taxonomy discovery processes in the \textbf{{point embedding space} (a-c)} and the \textbf{{box embedding space} (d-f)} of most existing HNTMs and the proposed BoxTM, respectively.}
\label{fig:point-vs-box}
\end{figure}

Considering the above challenges, we propose to learn topic taxonomy in the box embedding space \cite{vilnis2018prob} and develop a Box embedding-based Topic Model (BoxTM)\footnote{The source code of our model is available in public at: \url{https://github.com/luyy9apples/BoxTM}.} following the framework of NVI. Figure \ref{fig:point-vs-box} shows the differences of the topic taxonomy discovery processes in the point embedding space and the box embedding space, which are adopted by most existing HNTMs and our BoxTM, respectively. And the topic taxonomy discovery process in the hyperbolic embedding space is similar to that in the point embedding space. Specifically, BoxTM represents a topic or word as a hyperrectangle instead of a point, whose volume is proportional to the size of its semantic scope. In other words, the box embedding of a general topic covers a relatively larger region than that of a specific topic. Additionally, we conduct recursive clustering on the box embeddings of the lower-level topics to extract the upper-level topics. This approach leverages the connection between descendant topics to precisely capture the semantics of the upper-level topics, which can address the topic collapse problem caused by unguided upper-level topic mining. Intuitively, we employ symmetry and asymmetry distance metrics defined in the box embedding space respectively to capture similarity and hierarchy relations among topics. In summary, the main contribution of this paper is as follows:

\begin{itemize}
\item{We propose representing topics and words as box embeddings to capture their semantic scopes and accurately infer the hierarchical relations among these topics.}
\item{We propose to conduct recursive clustering on leaf topics to mine upper-level topics, which is an interpretable and effective way to capture the semantics of upper-level topics.}
\item{We conduct intrinsic evaluation, extrinsic evaluation, human evaluation, and qualitative analysis to validate the effectiveness of our model compared to state-of-the-art baselines.}
\end{itemize}

\section{Related Work}
\subsection{Document Generation-based Methods}
The classic topic model, i.e., LDA \cite{blei2003latent}, uses a document generative process under the framework of probabilistic graphical models to extract flat topics. As an extension of LDA to topic taxonomy discovery, a series of hierarchical topic models has been proposed, such as nCRP \cite{blei2003hierarchical} and rCRP \cite{kim2012modeling}. Despite their popularity, they suffer from high complexity of posterior inference. Recently, HNTMs \cite{isonuma2020tree,chen2021hierarchical}, based on NVI and deep generative model, are developed to tackle this problem. 

Inspired by the Embedded Topic Model (ETM) \cite{dieng2020topic}, nTSNTM \cite{chen2021tree} and SawETM \cite{duan2021sawtooth} project topics and words into the same Euclidean embedding space and construct topic taxonomy via the symmetric distances between topic and word points. Due to the advantage of hyperbolic space in modeling tree-structured data \cite{nickel2017poincare}, HyperMiner \cite{shi2022hyperminer} adopts a hyperbolic embedding space to discover topic taxonomy. However, HyperMiner still uses the symmetric distance metric (i.e., dot product) to infer the complex relations among topics and randomly initializes topic embeddings, following prior HNTMs. Such approximation of asymmetric relations and ``cold start'' of embedding learning result in a risk of top-level topics collapsing into meaningless common words. To alleviate the latter problem, C-HNTM \cite{wang2023hierarchical} attempts to learn topics of different levels using different semantic patterns. Specifically, C-HNTM learns level-2 topics by clustering on word embeddings, and it adopts ETM to mine leaf topics. Unfortunately, C-HNTM lacks the flexibility to learn topic taxonomies of different depths.

\subsection{Clustering-based Methods}
Since pre-trained embedding models \cite{devlin2019bert,pennington2014glove} have boosted the performance of many text mining tasks in recent years, a branch of research attempts to mine flat \cite{sia2020tired,meng2022topic} or hierarchical topics \cite{zhang2018taxogen,grootendorst2022bertopic} from high-quality embedding spaces directly. As a representative clustering-based method, TaxoGen \cite{zhang2018taxogen} conducts hierarchical clustering to group similar words into clusters (topics) and split coarse clusters (topics) into specific ones. Besides, it ranks the importance of each word to its topic by some manually designed metrics, such as the symmetric distance between a word and its cluster centroid. Importantly, most clustering-based methods train word embedding spaces on local contexts, which enables them to capture accurate semantics of words but hinders them from getting high-quality topics, because the boundaries between clusters are blurred in such delicate embedding spaces. Regardless, since topics are semantic summaries of corpora, global semantic information is more critical for topic mining compared to local contexts. However, clustering-based methods have trouble in utilizing the global statistics of word occurrences effectively. For example, both BERTopic \cite{grootendorst2022bertopic} and TaxoGen \cite{zhang2018taxogen} simply apply TF-IDF information as weights for topic keyword ranking.

\subsection{Supervised Methods}
Apart from self-supervised topic taxonomy discovery, another line of research tries to adopt a word-level knowledge graph \cite{lee2022taxocom,meng2020hierarchical} or manually built topic hierarchy \cite{duan2021topicnet} as the ``framework'' of the topic taxonomy. As a representative method of supervised HNTMs, TopicNet \cite{duan2021topicnet} adopts prior knowledge from WordNet \cite{miller1995wordnet}. Specifically, TopicNet discovers each topic and each topic hierarchical relation guided by a seed word and the hypernym-hyponym relation between seed words, respectively. Similarly, a clustering-based method named TaxoCom \cite{lee2022taxocom} uses manually defined seed words as centers of topic clusters. Unfortunately, there may be a semantic gap between the general knowledge graph and the target corpus, and it's difficult and costly to determine a complete topic hierarchy manually. Therefore, self-supervised topic taxonomy discovery is more flexible and versatile, since it does not rely on prior knowledge.

\section{Background Knowledge}

As a representative geometric embedding technology, the box embedding method represents a word or topic as a box (i.e., axis-aligned hyperrectangle) instead of a point in the traditional Euclidean embedding method. With extra degrees of freedom, box embeddings can capture semantic scopes and asymmetric relations of objects \cite{vilnis2018prob,li2019smoothing,dasgupta2020improving}.

\begin{definition}[\textit{box embedding}]
A $D$-dimensional box is determined by its minimum and maximum coordinates in each axis, parameterized by a pair of vectors $(\boldsymbol{x}_m,\boldsymbol{x}_M)$, where $\boldsymbol{x}_m,\boldsymbol{x}_M\in	\left[ 0,1 \right]^D$ and $\boldsymbol{x}_{m,i}\leq \boldsymbol{x}_{M,i}$, for $\forall i\in\{1\dots D\}$.
\end{definition}

\begin{definition}[\textit{box operations}]
Let $\mathrm{Box}(A):=(\boldsymbol{x}^A_m,\boldsymbol{x}^A_M),\mathrm{Box}(B):=(\boldsymbol{x}^B_m,\boldsymbol{x}^B_M)$ denote box embeddings of objects $A$ and $B$, respectively. The basic box operations are defined as follows:

\begin{subdefinition} (\textit{volume})\textbf{.}
The volume of $\mathrm{Box}(A)$ is defined as $\mathrm{Vol}(\mathrm{Box}(A)):=\prod_{i=1}^D (\boldsymbol{x}^A_{M,i}-\boldsymbol{x}^A_{m,i})$.
\end{subdefinition}

\begin{subdefinition} (\textit{intersection})\textbf{.}
If there is an overlap between $\mathrm{Box}(A)$ and $\mathrm{Box}(B)$, their intersection box is defined to be $\mathrm{Box}(A)\wedge \mathrm{Box}(B):=(\max (\boldsymbol{x}_m^{A},\boldsymbol{x}_m^{B}), \min (\boldsymbol{x}_M^{A},\boldsymbol{x}_M^{B}))$; otherwise, it is defined to be $\mathrm{Box}(A)\wedge \mathrm{Box}(B):=\perp$.
\end{subdefinition}

\begin{subdefinition} (\textit{union})\textbf{.}
The union box of $\mathrm{Box}(A)$ and $\mathrm{Box}(B)$ is defined as $\mathrm{Box}(A)\vee \mathrm{Box}(B):=(\min (\boldsymbol{x}_m^{A},\boldsymbol{x}_m^{B}), \max (\boldsymbol{x}_M^{A},\boldsymbol{x}_M^{B}))$.
\end{subdefinition}

\end{definition}

Note that box embeddings are closed under the intersection and union operations. For simplicity, the base box operations are described above, while in practice we adopt the gumbel version that is more stable for training \cite{dasgupta2020improving}.

In this work, we consider the volume of a topic or word box as its size of semantic scope, i.e., a more general concept covers a larger region in the latent semantic space. The union box of topics and words is a generalization of their semantics. For the symmetric affinity, denoted as $R_1$, there is $\forall A,B: A R_1 B \Rightarrow B R_1 A$. We estimate $R_1$ with the volume of the intersection between topic and word boxes ($\mathrm{R}_s$), which is defined as follows:
\begin{equation}
\label{eq:symmetric-relation}
\mathrm{R}_s(A,B) = \mathrm{Vol}\left(\mathrm{Box}(A)\wedge \mathrm{Box}(B)\right).
\end{equation}

Accordingly, we have $\mathrm{R}_s(A,B)=\mathrm{R}_s(B,A)$. To mitigate the bias towards large boxes, we can regularize the $\mathrm{R}_s(A,B)$ metric through division by $\mathrm{Vol}\left(\mathrm{Box}(A)\right)\cdot \mathrm{Vol}\left(\mathrm{Box}(B)\right)$ in practice.

For the asymmetric hierarchical relation between topics of adjacent levels, denoted as $R_2$, there is $\forall t^i,t^j\in\mathcal{T}:t^i R_2 t^j \Rightarrow \neg t^j R_2 t^i$, which means ``if $t^i$ { is a sub-topic of } $t^j$, then $t^j$ { is NOT a sub-topic of } $t^i$''. We reflect $R_2$ by the ratio of the volume of their intersection box to the upper-level topic box ($\mathrm{R}_a$), that is,
\begin{equation}
\label{eq:asymmetric-relation}
\mathrm{R}_a\left(t_k^i \big| t_{k+1}^j\right) = \frac{\mathrm{Vol}\left(\mathrm{Box}(t_k^i)\wedge \mathrm{Box}(t_{k+1}^j)\right)}{\mathrm{Vol}\left(\mathrm{Box}(t_{k+1}^j)\right)},
\end{equation}
where $t_k^i\in\mathcal{T}_k$ and $t_{k+1}^j\in\mathcal{T}_{k+1}$ denote topics of the $k$-th and ($k$+1)-th level, respectively. Unlike $\mathrm{R}_s(\cdot,\cdot)$,  $\mathrm{R}_a(\cdot|\cdot)$ has the property that $\mathrm{R}_a(A\big|B)=\mathrm{R}_a(B\big|A)\ne 0 \text{ iff. } \mathrm{Vol}\left(\mathrm{Box}(A)\right)=\mathrm{Vol}\left(\mathrm{Box}(B)\right)$. Thus $\mathrm{R}_a(\cdot|\cdot)$ can better model the hierarchical relation that is asymmetric.

\textbf{Discussion of box embeddings for taxonomy learning:} Most of the previous works \cite{vilnis2018prob,DBLP:conf/coling/LeesWZKC20,dasgupta2020improving} learn box embeddings of pre-defined entities or words for taxonomy completion in a supervised manner. For instance, \citet{vilnis2018prob} first proposed to train box embeddings for words on the incomplete ontology, in order to infer missing hypernym relations. Unlike these supervised methods, this paper aims at self-supervised topic taxonomy construction from unstructured text via box embeddings. This research problem poses new challenges for box embedding learning. Accordingly, we propose a recursive clustering algorithm for self-supervised box embedding learning, which is integrated with a VAE framework to provide an efficient solution for topic taxonomy construction based on box embeddings.

\section{Proposed Method}

In this section, we introduce the proposed BoxTM in detail. Firstly, we propose the box embedding-based document generative process in Section \ref{sec:boxtm}, which is the main framework of BoxTM. In general, BoxTM infers topic distributions via the symmetric affinities and semantic scopes of topics and words in the box embedding space. Additionally, the hierarchical relations are modeled by the values of the asymmetric metric between topic boxes. Subsequently, we introduce more detailed designs of BoxTM, including a novel workflow of recursive topic clustering for upper-level topic mining (Section \ref{sec:recur-clus}) and two self-training tasks for modeling the semantic scopes of words and topics better (Section \ref{sec:ssm}). Finally, we introduce the learning strategy of BoxTM in Section \ref{sec:learning}.

\subsection{Document Generative Process}
\label{sec:boxtm}

BoxTM holds the assumption that a document is generated by any topics in the topic taxonomy and adopts a bottom-up hierarchical topic discovery method following \citet{chen2021tree}. For NVI, BoxTM adopts a classic Variational AutoEncoder (VAE) with a logistic normal distribution $\mathcal{LN}(\mathbf{0},\boldsymbol{I})$ \cite{atchison1980logistic} as the prior of topic proportion. A VAE consists of an encoder that learns hierarchical topic proportions given document representations and  a decoder that reconstructs documents based on hierarchical topic proportions and topic distributions. Figure \ref{fig:boxtm} shows the main framework of BoxTM. 

Given a corpus $\mathcal{D}$ and a vocabulary $\mathcal{V}$, BoxTM firstly encodes the TF-IDF representation $\boldsymbol{d}\in\mathbb{R}^{|\mathcal{V}|}$ of each document into a latent distribution, from which the latent feature $\boldsymbol{z}$ is sampled. After transforming $\boldsymbol{z}$ to acquire the leaf topic proportion $\boldsymbol{\pi}_1$, we infer upper-level topic proportions $\{\boldsymbol{\pi}_{>1}\}$ based on the asymmetric relations $\{\Theta_k\}$ of topics in the box embedding space. Specifically, $\Theta_k\in\mathbb{R}^{|\mathcal{T}_k|\times|\mathcal{T}_{k+1}|}$ between level-$k$ topics $\mathcal{T}_k$ and the upper-level topics $\mathcal{T}_{k+1}$ are estimated by the asymmetric metric $\mathrm{R}_a\left(\cdot|\cdot\right)$, i.e.,
\begin{equation}
\label{eq:hier-topic}
    \Theta_k^{ij}=\log\mathrm{R}_a\left(t_{k}^i \big| t_{k+1}^j\right),
\end{equation}
where  $t_{k}^i\in\mathcal{T}_k$ and $t_{k+1}^j\in\mathcal{T}_{k+1}$. The encoding process of BoxTM is defined as follows:
\begin{align}
\label{eq:encode}
    \boldsymbol{h} &= f_{\boldsymbol{h}}(\boldsymbol{d}), \\
    \boldsymbol{z} &\sim \mathcal{N}\left(f_{\boldsymbol{\mu}}(\boldsymbol{h}),f_{\boldsymbol{\sigma}}(\boldsymbol{h})\right),\\ 
    \boldsymbol{\pi}_1 &= \mathrm{Softmax}\left(f_{\boldsymbol{\pi}}(\boldsymbol{z})\right),\\
    \label{eq:topic-word-dir}
    \boldsymbol{\pi}_{k+1} &= \mathrm{Softmax}\left(\boldsymbol{\pi}_k\Theta_k\right),
\end{align}
where $f_{\boldsymbol{h}}(\cdot),f_{\boldsymbol{\mu}}(\cdot)$, $f_{\boldsymbol{\sigma}}(\cdot)$, and $f_{\boldsymbol{\pi}}(\cdot)$ are feedforward neural networks. As the sampling process for the latent feature $\boldsymbol{z}$ is not differentiable, we adopt the reparameterization trick \cite{rezende2014stochastic} to make the gradient descent possible. Specifically, the sampled feature $\boldsymbol{z}$ can be expressed by a standard normal distribution, i.e., $\boldsymbol{z} = f_{\boldsymbol{\mu}}(\boldsymbol{h}) + \epsilon\cdot f_{\boldsymbol{\sigma}}(\boldsymbol{h}), \epsilon \sim \mathcal{N}(\boldsymbol{0}, \boldsymbol{I})$.

\begin{figure}[!t]
\centering
\includegraphics[width=\linewidth]{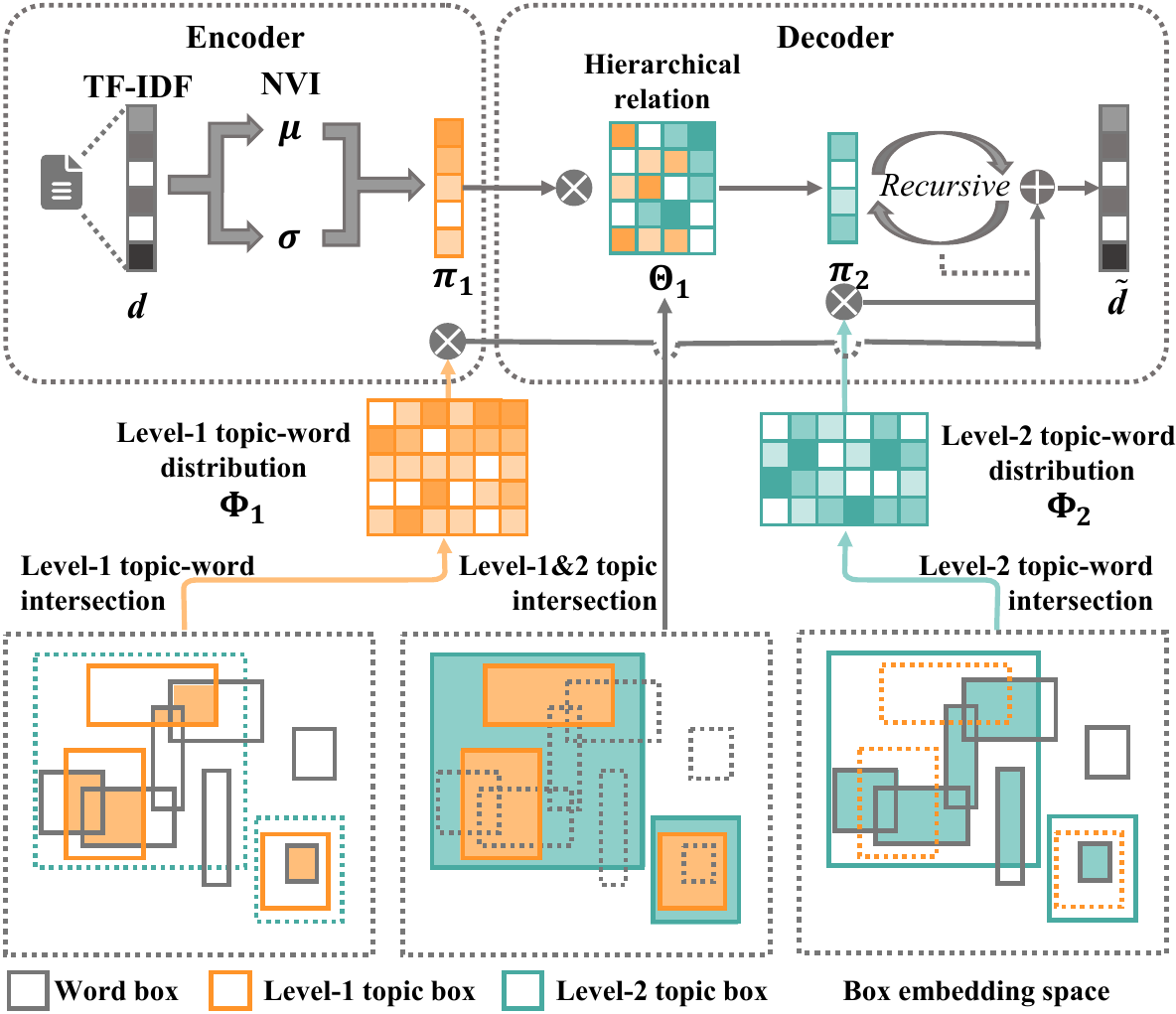}
\caption{The main framework of BoxTM.}
\label{fig:boxtm}
\end{figure}

For the decoding process of BoxTM, we apply normalization before document reconstruction to enhance the generation power of weak topic levels \cite{hung2019scops}, which is defined as follows:
\begin{equation}
\label{eq:decode}
    \boldsymbol{\tilde{d}} = \sum\limits_{k=1}^{K} (\boldsymbol{\pi}_k \cdot \Phi_k) \circ \mathrm{CV}_{\Phi_k} / Z_k, 
\end{equation}
where $K$ is the depth of the topic taxonomy and $\circ$ denotes the element-wise multiplication. $\Phi_k\in[0,1]^{|\mathcal{T}_k|\times|\mathcal{V}|}$ is topic-word distributions of the $k$-th level and $Z_k=||(\boldsymbol{\pi}_k \cdot \Phi_k) \circ \mathrm{CV}_{\Phi_k}||_2$ is a 2-norm term. To weaken the impact of common words on document generation, we adopt the Coefficient of Variation (CV) \cite{brown1998coefficient} to sharpen all topic-word distributions $\{\Phi_k\}$. Specifically, the $j$-th element of $\mathrm{CV}_{\Phi_k}\in\mathbb{R}^{|\mathcal{V}|}$ is the ratio of the standard deviation to the mean of the $j$-th column in $\Phi_k$, which is defined by $\mathrm{CV}_{\Phi_k}^j = \sigma(\Phi_k^{:,j})/\mu(\Phi_k^{:,j})$.

Notably, BoxTM infers topic-word distributions over the vocabulary $\mathcal{V}$ via the normalized symmetric affinity between topic and word boxes. For the $i$-th topic $t_k^{i}$ at level-$k$ and the $j$-th word $w_j$ in $\mathcal{V}$,
\begin{equation}
\label{eq:topic-distribution}
    \Phi_{k}^{ij} = \mathrm{Softmax}\left(\log\frac{\mathrm{R}_s(t_k^i,w_j)}{\mathrm{Vol}(t_k^i)\cdot\mathrm{Vol}(w_j)}\right),
\end{equation}
which enables abstract topics to bias toward general words, and vice versa. 

In summary, we describe the document generative process of BoxTM as follows: 

\noindent $\triangleright$ \textit{For global topics,  $k\in\{1,\dots,K$-1$\}$:}

\noindent 1. Infer the hierarchical relations between level-$k$ and level-($k$+1) topics $\Theta_k$ by Eq. (\ref{eq:hier-topic}).

\noindent 2. Infer the topic-word distribution $\Phi_k$ by Eq. (\ref{eq:topic-distribution}).

\noindent $\triangleright$ \textit{For each document:}

\noindent 1. Draw the leaf topic proportion $\boldsymbol{\pi}_1\sim\mathcal{LN}(\mathbf{0},\boldsymbol{I})$.

\noindent 2. Infer the upper-level topic proportion $\boldsymbol{\pi}_{k+1}$ by Eq. (\ref{eq:topic-word-dir}), for level $k\in\{1,\dots,K$-1$\}$.

\noindent 3. For each word $w_j$ in the document:

a. Draw topic level $k\sim\textbf{Uniform}(K)$.

b. Draw topic assignment $t_k^{i}\sim\textbf{Cat}(\boldsymbol{\pi}_k)$.

c. Draw word $\hat{w}_{j}\sim\textbf{Cat}(\Phi_k^{i,:})$.

\subsection{Recursive Topic Clustering}
\label{sec:recur-clus}
Unlike most HNTMs that randomly initialize embeddings of topics in different abstraction levels, BoxTM conducts recursive clustering on topic boxes to learn upper-level topics. Notably, such a method can alleviate the problem of topic collapse, since the upper-level topic mining is guided by the correlation between lower-level topics. For the selection of clustering algorithms, we adopt the Affinity Propagation (AP) \cite{frey2007clustering} algorithm for its flexibility and interpretability\footnote{Compared to the AP algorithm, centroid-based methods such as k-means++ \cite{arthur2007kmeans} cannot accommodate non-flat geometries like the box embedding space, while density-based DBSCAN \cite{ester1996dbscan} is vulnerable to the setting of hyperparameters.}.

BoxTM constructs a topic affinity graph for topics at each level, where topic nodes are connected if their boxes overlap. However, the direct correlation between topics may be sparse in the box embedding space due to the diversity of topics, i.e., $\mathrm{Vol}(\mathrm{Box}(t_k^{i})\wedge\mathrm{Box}(t_k^{j}))\rightarrow 0$, $\forall t_k^{i},t_k^{j}\in\mathcal{T}_k$. To address this, we expand the semantic scope of each topic by merging the information of its keyword boxes. The box embedding of the processed $i$-th topic $\tilde{t}_k^i$ at level-$k$ is defined as follows:
\begin{equation}
\label{eq:expand-box}
    \mathrm{Box}(\tilde{t}_k^{i}):= \left[\vee_{w\in \mathcal{W}_k^{i}}\mathrm{Box}(w)\right] \vee \mathrm{Box}(t_k^{i}),
\end{equation}
where $\mathcal{W}_k^{i}=\{w_j|\arg\max_j\Phi_k^{ij}\}$ with $\left|\mathcal{W}_k^{i}\right|=n$ denotes the set of top-$n$ ($n$ = 5 in our experiments) representative words of topic $t_k^{i}$. Next, the affinity between topics is measured by the value of the asymmetric metric $\mathrm{R}_a(\cdot|\cdot)$ instead of the symmetric similarity metric $\mathrm{R}_s(\cdot,\cdot)$, because $\mathrm{R}_a(\cdot|\cdot)$ can weaken the influence of hub topics in clustering and prevent over-smoothing. Formally, the affinity matrix $\mathcal{A}_k\in\mathbb{R}^{|\mathcal{T}_k|\times|\mathcal{T}_k|}$ is defined by
\begin{equation}
\label{eq:topic-aff}
    \mathcal{A}_k^{ij} = \left\{
\begin{aligned}
& \log\mathrm{R}_a(\tilde{t}_k^{j}\big|\tilde{t}_k^{i}) &, i\ne j;\\
& 0 &, i=j.
\end{aligned}
\right.
\end{equation}

Later, the union of topic boxes in each cluster is adopted as a reasonable initialization of an upper-level topic. To reduce the impact of outliers in clustering, we propose a soft union operation $\vee_{\dagger}$, which is defined as follows:
\begin{equation}
\label{eq:soft-union}
\begin{aligned}
    & \mathrm{Box}\left( t_{k+1}^i \right) := (x_m^{i},x_M^{i}) = \vee_{\dagger t\in \mathcal{C}_{k}^i}\mathrm{Box}(\tilde{t}),\\
    & x_m^{i} = \mu(\{x_m^{t}\}_{t\in\mathcal{C}_{k}^i}),x_M^{i}=\mu(\{x_M^{t}\}_{t\in\mathcal{C}_{k}^i}),
\end{aligned}
\end{equation}
where $\mathcal{C}_{k}^i$ is the $i$-th topic cluster of the $k$-th level and $\mu(\cdot)$ is the mean operation. Besides, $\mathrm{Box}\left( t_{k+1}^i \right)$ is the reinitialized box embedding for the upper-level topic $t_{k+1}^i$. Then BoxTM infers the hierarchical relations $\Theta_k$ between level-$k$ and level-($k$+1) topics based on their box embeddings. For each topic $t_{k}^i\in\mathcal{T}_k$ at the $k$-th level, its most relevant topic at the upper level is adopted as its parent topic $t_{p}^i\in\mathcal{T}_{k+1}$. Formally, we have 
\begin{equation}
\label{eq:topic-hier-relation}
    t_{p}^i := t_{k+1}^j = \arg\max_j\Theta_k^{ij}.
\end{equation}

After conducting ($K$-1) times of topic clustering recursively, BoxTM can mine topics of $K$ levels in a bottom-up manner.

\subsection{Semantic Scope Modeling}
\label{sec:ssm}
The effectiveness of our box embedding-based document generative process with recursive topic clustering is based on an important premise that box embeddings can accurately model the semantic scopes of words and topics. Here we propose two self-supervised tasks  by means of word-level and topic-level constraints for semantic scope modeling.

\subsubsection{Word-level Constraint} Importantly, the semantic scope of each word consists of its \textbf{\textit{abstraction level}} and \textbf{\textit{semantics}}, which correspond to the \textbf{\textit{volume}} and \textbf{\textit{position}} of its box, respectively. Inspired by GloVe \cite{pennington2014glove}, we propose to encode the (co-)occurrence patterns of words into word boxes. 

Our key insight is that the marginal probability $P(w_j)$ of word $w_j$ reveals its abstraction level. Besides, as the distributional hypothesis states that similar words $w_i$ and $w_i'$ tend to co-occur with the same word $w_j$, the joint probability $P(w_i,w_j)$ may reflect the correlation between the semantics of $w_i$ and $w_j$. In practice, the joint and marginal probabilities can be estimated by $P(w_i,w_j)\sim X_{ij}$ and $P(w_j)\sim X_j$, where $X_{ij}$ is the co-occurrence time of $w_i$ and $w_j$ in the corpus, and $X_j=\sum_{w_n\in\mathcal{V}}X_{jn}$. Integrating these patterns, we propose that the values of the asymmetric metric $\mathrm{R}_a(w_i|w_j)$ in the box embedding space should be consistent with the conditional probability $P_{i|j}=P(w_i|w_j)=X_{ij}/X_j$.

For the word-level constraint of semantic scope modeling, the Mean-Square Error (MSE) loss is a straightforward selection, i.e., $\mathcal{L}_{CO}=\left\| \mathrm{R}_a(w_i|w_j) - P_{i|j} \right\|_2^2$. However, the MSE loss strongly restricts the absolute volumes of word boxes, which is difficult for training. Therefore, we adopt the cross-entropy loss $H(\cdot,\cdot)$ to constrain the relative volumes of word boxes among a randomly sampled batch $\mathcal{B}=\{(w_i,w_j)| P_{i|j}>0 \}$. Formally, we denote the box volume distribution as $q_{\mathrm{Box}}(w_i,w_j)\sim\mathrm{R}_a(w_i|w_j)$ and the co-occurrence pattern distribution as $p_{CO}(w_i,w_j)\sim P_{i|j}$. Then the loss function is defined by
\begin{equation}
\label{eq:L-CO}
\begin{aligned}
    \mathcal{L}_{CO} &= H(p_{CO},q_{\mathrm{Box}}) \\
    &=-\sum\limits_{(w_i,w_j)\in\mathcal{B}}p_{CO}(w_i,w_j)\log q_{\mathrm{Box}}(w_i,w_j).
\end{aligned}
\end{equation}

\subsubsection{Topic-level Constraint}
In a reasonable topic taxonomy $\mathcal{S}$, the semantic scope of a parent topic $t_p$ should cover that of its child topic $t_c$ \cite{viegas2020cluhtm}. In other words, the box embedding of $t_p$ should entail that of $t_c$. Intuitively, we can define the following loss to maximize the score of asymmetric correlation metric between $t_p$ and $t_c$:
\begin{equation}
\begin{aligned}
\label{eq:l-ra}
    \mathcal{L}_{HT} &= -\sum_{(t_p,t_c)\in\mathcal{S}} \log\mathrm{R}_a(t_c|t_p) \\
    &= -\sum_{(t_p,t_c)\in\mathcal{S}} \log\mathrm{R}_s(t_c,t_p) - \log\mathrm{Vol}(t_p),
\end{aligned}
\end{equation}
where the first term $\mathrm{R}_s(t_c,t_p)$ regularizes the semantic coherence between $t_p$ and $t_c$. However, the second term of the above definition may lead to a trivial solution that all topic boxes collapse to points, i.e., $\mathrm{Vol}(t)\rightarrow 0$ and then $\mathrm{R}_s(t_c,t_p)\rightarrow 0$, $\forall t,t_c,t_p$. To avoid this problem, we replace the second term with a max-margin objective, which makes the box of $t_p$ larger than that of $t_c$ by at least the margin $m$. So $\mathcal{L}_{HT}$ is redefined as follows:
\begin{equation}
\begin{aligned}
\label{eq:L-HT}
    \mathcal{L}_{HT} = &-\sum_{(t_p,t_c)\in\mathcal{S}} \log\mathrm{R}_s(t_c,t_p) \\
    &- \max\left[0,m-\log\mathrm{Vol}(t_p)+\log\mathrm{Vol}(t_c)\right].
\end{aligned}
\end{equation}

\begin{algorithm}[!t]
\caption{The $i$-th epoch of training}\label{alg:train}

\hspace*{0.02in}{\textbf{Input:} The corpus $\mathcal{D}$ and its vocabulary $\mathcal{V}$; The word and topic box embeddings $W$ and $\{T_k\}$; The topic taxonomy $\mathcal{S}$ after prior epoch; The threshold $\gamma$ for early stop.}\\
\hspace*{0.02in}{\textbf{Output:} Updated word and topic box embeddings $\tilde{W}$ and $\{\tilde{T}_k\}$; Updated topic taxonomy $\tilde{\mathcal{S}}$.}

\begin{algorithmic}[1]
\STATE \textbf{if} $i<\gamma$ \textbf{then}
\STATE \hspace{0.2cm} $\tilde{\mathcal{S}},\{\tilde{T}_k\} \leftarrow$ \textsc{RecurClus}$(W,T_1,K)$
\STATE \textbf{else} $\tilde{\mathcal{S}},\{\tilde{T}_k\} \leftarrow \mathcal{S},\{T_k\}$
\STATE \textbf{for} each batch $\mathcal{B}\subset \mathcal{D}$ \textbf{do}
\STATE \hspace{0.2cm} Infer hierarchical relations $\Theta$ by Eq. (\ref{eq:hier-topic}).
\STATE \hspace{0.2cm} Infer topic-word distributions $\Phi$ by Eq. (\ref{eq:topic-distribution}).
\STATE \hspace{0.2cm} \textbf{for} each document $\boldsymbol{d}\in \mathcal{D}$ \textbf{do}
\STATE \hspace{0.4cm} Draw topic proportions $\{\boldsymbol{\pi}_k\} \leftarrow$ \textsc{Encode}$(\boldsymbol{d},\Theta)$.
\STATE \hspace{0.4cm} Reconstruct document $\boldsymbol{\tilde{d}} \leftarrow $ \textsc{Decode}$(\{\boldsymbol{\pi}_k\},\Phi)$.
\STATE \hspace{0.2cm} Compute loss $\mathcal{L} = \mathcal{L}_{ELBO} + \mathcal{L}_{Box}$.
\STATE \hspace{0.2cm} Update $\tilde{W}$ and $\{\tilde{T}_k\}$ by minimizing $\mathcal{L}$.
\STATE \hspace{0.2cm} Update $\tilde{\mathcal{S}}$ based on $\{\tilde{T}_k\}$ by Eq. (\ref{eq:topic-hier-relation}).
\STATE \textbf{return} $\tilde{W}$, $\{\tilde{T}_k\}, \tilde{\mathcal{S}}$
\end{algorithmic}
\end{algorithm}

\subsection{Learning Strategy}
\label{sec:learning}
Similar to the training objective of VAEs, the main loss of BoxTM is to maximize the Evidence Lower BOund (ELBO). Specifically, the ELBO loss of BoxTM is defined by
\begin{equation}
\label{eq:elbo}
\begin{aligned}
    \mathcal{L}_{ELBO} =& \mathbb{E}_{\boldsymbol{\pi}_1\sim q_{\boldsymbol{d}}} {\log p(\boldsymbol{d}| \{\boldsymbol{\pi}_k\} ,\{\Phi_k\})} \\
    &- D_{KL}\left[ q_{\boldsymbol{d}}(\boldsymbol{\pi}_1) || p(\boldsymbol{\pi}_1) \right],
\end{aligned}
\end{equation}
which balances between maximising the expected log-likelihood (the first term) and minimising the KL divergence (the second term) of the variational distribution $q_{\boldsymbol{d}}(\boldsymbol{\pi}_1):=\mathcal{N}(f_\mu(\boldsymbol{d}),f_\sigma(\boldsymbol{d}))$ and the prior distribution $p(\boldsymbol{\pi}_1):=\mathcal{N}(\boldsymbol{0},\boldsymbol{I})$.

For modeling the semantic scopes of words and topics, we propose two constraints in Section \ref{sec:ssm}. Accordingly, we define the regularization loss by
\begin{equation}
\label{eq:box-reg-loss}
\mathcal{L}_{Box} = \alpha\cdot\mathcal{L}_{CO} + \beta\cdot\mathcal{L}_{HT},
\end{equation}
where $\alpha$ and $\beta$ are weights for these losses. And the overall loss function of BoxTM is defined by
\begin{equation}
\label{eq:final-loss}
\mathcal{L}=\mathcal{L}_{ELBO}+\mathcal{L}_{Box}.
\end{equation}

Then we adopt the Adam optimizer to update the network parameters of the encoder and box embeddings of topics and words. Based on the updated topic boxes, we perform a correction for the topic taxonomy using Eq. (\ref{eq:topic-hier-relation}). The training workflow of BoxTM is shown in Algorithm \ref{alg:train}. Intuitively, topic boxes overlap less along with the training to capture diverse semantics, which limits the effectiveness of our recursive clustering module at the late phase of training. To tackle this problem, we use the early stopping trick that stops recursive clustering after the $\gamma$-th iteration. In the following experiments, $\gamma$ is set to 100.

\section{Experiments}
\subsection{Experimental Settings}
\subsubsection{Datasets} 
We conduct comprehensive evaluations on three benchmark datasets with latent topic hierarchies: (1) \textbf{20news}\footnote{\url{http://qwone.com/~jason/20Newsgroups/}}: A corpus consists of 20 newsgroups \cite{song2014on}. (2) \textbf{NYT}\footnote{\url{http://developer.nytimes.com/}}: A set of news articles from the New York Times, which are categorized into 25 classes. (3) \textbf{arXiv}\footnote{\url{https://arxiv.org/}}: A set of paper abstracts covering 53 classes from arXiv website. The latter two datasets are collected by \citet{meng2019weakly}. Table \ref{tab:dataset} shows the statistics of all datasets. After preprocessing of removing stopwords and low-frequency words, we split documents into a training set and a testing set with the ratio of 6:4. In addition, we adopt 20\% of documents in the training set as a validation set.

\subsubsection{Baselines}
We compare our model with state-of-the-art topic taxonomy discovery models based on different frameworks, including document generation-based methods of \textbf{nTSNTM}\footnote{\url{https://github.com/hostnlp/nTSNTM}} \cite{chen2021tree}, \textbf{SawETM}\footnote{\url{https://github.com/BoChenGroup/SawETM}} \cite{duan2021sawtooth}, \textbf{HyperMiner}\footnote{\url{https://github.com/NoviceStone/HyperMiner}} \cite{shi2022hyperminer}, and \textbf{C-HNTM}\footnote{\url{https://github.com/Jladygoogoo/C-HNTM}} \cite{wang2023hierarchical}, as well as a clustering-based method of \textbf{TaxoGen}\footnote{\url{https://github.com/franticnerd/taxogen}} \cite{zhang2018taxogen}. Notably, HyperMiner adopts the hyperbolic embedding space, and the others hold the Euclidean embedding space assumption.

\subsubsection{Hyperparameter settings}
The maximum depth of the topic taxonomy is set to 3 for the 20news and NYT datasets following \citet{chen2021tree}. To evaluate the flexibility of BoxTM and baseline models, the maximum depth for the large dataset arXiv is set to 5. Besides, the maximum number of leaf topics $|\mathcal{T}_1|^{\max}$ of nTSNTM is 200 following the setting in its paper, which can get a reasonable number of topics adaptively based on the stick-breaking process. According to the number of active topics obtained by nTSNTM, $|\mathcal{T}_1|^{\max}$ of BoxTM and the other HNTMs is set to 50/50/100 for three datasets, respectively. For TaxoGen, the maximum number of clusters is set to 5/5/3. The embedding dimension of BoxTM is set to 50 following \citet{vilnis2018prob}. Since box embeddings have 2 parameters per dimension, the embedding size of baselines are set to 100 for a fair comparison. 

Other hyperparameters of baselines take the optimal values reported in their papers. For BoxTM, the learning rate is 5e-3, the dimension of hidden layers is 256, and the max margin $m$ is set to 10. The weight of $\mathcal{L}_{HT}$ gradually increases to the maximum value ($\beta^{max}$ = 0.005) during training, when the constant weight of $\mathcal{L}_{CO}$ is set to 3.

\begin{table}[!t]
\caption{Statistics of datasets.}
\label{tab:dataset}
\centering
\resizebox{0.9\linewidth}{!}{
\begin{tabular}{|c|ccc|c|c|}
\hline
\multirow{2}{*}{\textbf{dataset}} & \multicolumn{3}{c|}{\textbf{\#document}} & \multirow{2}{*}{\textbf{\#word}} & \multirow{2}{*}{\textbf{\#class}} \\
 & \textbf{\#train} & \textbf{\#valid} & \textbf{\#test} &  &  \\
\hline \hline
20news & 9,007 & 2,251 & 7,487 & 1,838 & 20 \\
NYT & 6,279 & 1,569 & 5,233 & 8,171 & 25 \\
arXiv & 110,451 & 27,612 & 92,042 & 11,799 & 53 \\ 
\hline
\end{tabular}}
\end{table}

\subsection{Intrinsic Evaluation of Topic Taxonomy}
For a reasonable topic taxonomy, each topic is a set of closely coherent words and diverse from one another. Besides, keywords of a parent topic $t_p$ and its child topic $t_c$ are coherent but have different semantic abstraction levels. Thus we validate the quality of the topic taxonomy from the following perspectives: (1) \textbf{Topic Coherence (C)}: We adopt a classic metric NPMI \cite{lau2014machine} to quantify the coherence of mined topics. (2) \textbf{Topic Diversity (D)}: The widely-used TU \cite{nan2019topic} metric is for assessing the diversity among all topics, which is calculated by the number of unique keywords among all topics. (3) \textbf{Hierarchical Coherence (HC)}: We adopt the CLNPMI \cite{chen2021tree} metric to evaluate the hierarchical coherence between topics $t_p$ and $t_c$.

Because highly overlapping topics may cause inflated coherence scores, the product of NPMI and TU are used as an integrated metric (\textbf{C*D}) for a comprehensive validation \cite{dieng2020topic}. For the aforementioned metrics, we calculate the average of the scores of top-5, top-10, and top-15 topic words. Because the source code of nTSNTM and the algorithm of C-HNTM cannot adapt to topic taxonomy with more than 3 levels, their results on the arXiv dataset are not reported.

\begin{table*}[!t]
\caption{Intrinsic metric scores on three datasets.}
\label{tab:taxo-result}
\renewcommand{\arraystretch}{1.15}
\centering
\resizebox{0.9\linewidth}{!}{
\begin{tabular}{|c|cccc|cccc|cccc|}
\hline
 & \multicolumn{4}{c|}{\textbf{20news}} & \multicolumn{4}{c|}{\textbf{NYT}} & \multicolumn{4}{c|}{\textbf{arXiv}} \\
\textbf{model} & \textbf{C} & \textbf{D} & \textbf{C*D} & \textbf{HC} & \textbf{C} & \textbf{D} & \textbf{C*D} & \textbf{HC} & \textbf{C} & \textbf{D} & \textbf{C*D} & \textbf{HC} \\ 
\hline \hline
nTSNTM & 0.212 & 0.728 &	0.154 &	0.134 
 & 0.221 & 0.420 & 0.093 &	0.079 
 & - & - &	- &	- 
 \\ 
 SawETM & 0.221 & 0.404 &	0.089 &	0.098 
 & 0.228 & 0.476 &	0.109 &	0.084 
 & 0.134 & 0.256 &	0.034 &	0.047 
 \\
HyperMiner & 0.224 & 0.459 &	0.103 &	0.102
 & 0.231 & 0.500 &	0.115 &	0.101 
 & 0.142 & 0.382 &	0.054 &	0.050 
 \\
C-HNTM & 0.196 & 0.633 &	0.124 &	0.090 
 & 0.152 & 0.458 &	0.070 &	0.036 
 & - & - &	- &	- 
 \\ 
TaxoGen & 0.202 & \textbf{0.789} &	0.159 &	0.123 
 & 0.239 & \textbf{0.881} &	0.210 &	0.111 
 & 0.214 & \textbf{0.681} &	0.146 &	0.084 
 \\
\hline
BoxTM & \textbf{0.301} & 0.661 &	\textbf{0.199} &	\textbf{0.159}
& \textbf{0.409} & 0.648 &	\textbf{0.265} &	\textbf{0.177} 
& \textbf{0.257} & 0.672 &	\textbf{0.173} &	\textbf{0.113} 
\\
\hline
\end{tabular}}
\end{table*}

\begin{figure}[!t]
\centering
\includegraphics[width=\linewidth]{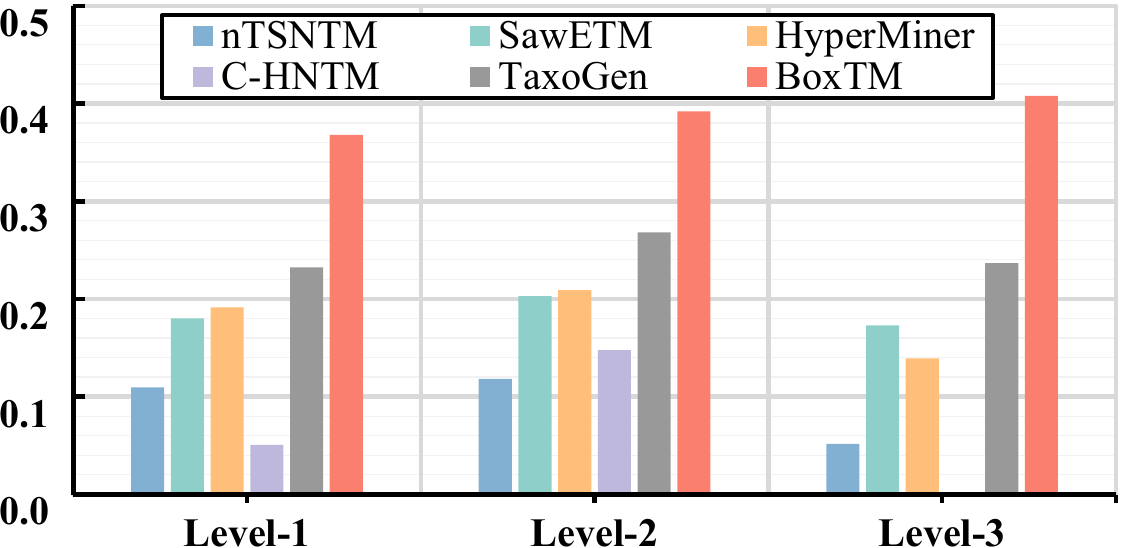}
\caption{The \textbf{C*D} scores at each level of BoxTM and baselines on NYT.}
\label{fig:nyt-level-tq}
\end{figure}

As shown in Table \ref{tab:taxo-result}, BoxTM achieves new state-of-the-art results on most metrics across three datasets, when HyperMiner using hyperbolic embeddings outperforms SawETM. These results validate the advantage of geometric (i.e., hyperbolic and box) embeddings on topic taxonomy discovery over traditional point embeddings. Compared to C-HNTM that performs poorly on the \textbf{HC} metric, the proposed recursive topic clustering module of BoxTM can effectively learn topics of different levels. While both SawETM and HyperMiner fail to learn a deep topic taxonomy on the arXiv dataset with massive documents, BoxTM remains outstanding performance on topic quality and hierarchical coherence. It validates that BoxTM not only has scalability for large-scale data but also has flexibility to learn topic taxonomies of different structures. In terms of the clustering-based method, TaxoGen obtains high scores of topic diversity (\textbf{D}), because each word only belongs to one topic at each level in its approach. However, it neglects the polysemy of some words, i.e., a word can be the keyword of different topics, which leads to its performance decline on topic coherence. For example, the word ``\textit{driver}'' could be the keyword of topics ``\textit{hardware}'' and ``\textit{motorcycles}''.

Furthermore, Figure \ref{fig:nyt-level-tq} illustrates the \textbf{C*D} scores at each level of BoxTM and baselines on the NYT dataset. Both coherence and diversity of the level-2 topics of all models have different degrees of improvement compared to leaf topics. However, most baselines fail to learn high-quality topics at the root level, that is, they encounter the topic collapse problem. And topics mined by BoxTM remain high-quality at all levels, due to the effectiveness of the proposed recursive topic clustering module.

\begin{table}[!t]
\caption{Extrinsic metric scores on three datasets.}
\label{tab:ex-result}
\renewcommand{\arraystretch}{1.15}
\centering
\resizebox{\linewidth}{!}{
\begin{tabular}{|c|cc|cc|cc|}
\hline
 & \multicolumn{2}{c|}{\textbf{20news}} & \multicolumn{2}{c|}{\textbf{NYT}} & \multicolumn{2}{c|}{\textbf{arXiv}} \\
\textbf{model} & \textbf{ARI} & \textbf{F$_\beta$} & \textbf{ARI} & \textbf{F$_\beta$} & \textbf{ARI} & \textbf{F$_\beta$}  \\ 
\hline \hline
nTSNTM & 0.081 &	0.133 
 &	0.389 &	0.448 
 & - & -
 \\ 
 SawETM & 0.074 &	0.123 
 & 0.452 &	0.494 
 & \textbf{0.151} &	\textbf{0.184}
 \\
HyperMiner & 0.075 &	0.127 
& 0.421 &	0.466 
& 0.115 &	0.151 
 \\
C-HNTM & 0.056 &	0.104 
 & 0.143 &	0.216 
 & - & - 
 \\ 
TaxoGen & 0.066 &	0.132 
 &	0.310 &	0.367 
 & 0.097 &	0.133 
 \\
\hline
BoxTM & \textbf{0.117} & \textbf{0.168}
& \textbf{0.541} & \textbf{0.577}
& 0.103 & 0.143 
\\
\hline
\end{tabular}}
\end{table}

\begin{figure*}[!t]
\centering
\includegraphics[width=5.5in]{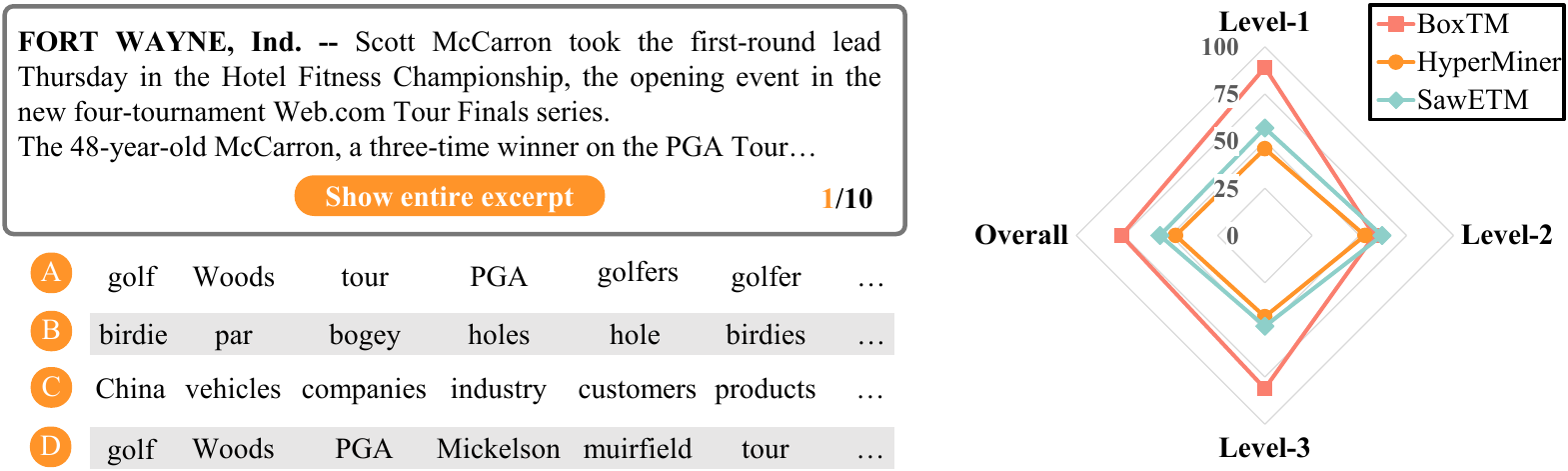}
\caption{Illustration of the human evaluation on the NYT dataset: an example of the topic intrusion task (left) and the average precision (\%) of our BoxTM and strong baselines (right).}
\label{fig:topic-intrusion}
\end{figure*}

\subsection{Extrinsic Evaluation of Topic Taxonomy}
As an important application scenario for topic taxonomy discovery, the tree structure and keywords of the mined topic taxonomy can serve as auxiliary knowledge to improve the performance of hierarchical text clustering \cite{lee2022taxocom}. Specifically, each topic is regarded as a cluster, characterized by its keywords. We utilize the topic structure and the top-15 keywords of all topics learned by our BoxTM and baseline models as the inputs of a hierarchical text clustering model named WeSHClass \cite{meng2019weakly}. For the evaluation metrics, we adopt two external criteria of clustering (i.e., \textbf{ARI} and \textbf{F$_\beta$}) using golden labels of documents \cite{steinbach2005cluster}.

Table \ref{tab:ex-result} shows the results of BoxTM and baseline models on the hierarchical text clustering task. Particularly, BoxTM and other HNTMs significantly outperform C-HNTM and TaxoGen that conduct clustering on word embeddings to mine topics, which reveals the limitation of latter methods in learning document-level semantics. Among HNTMs, BoxTM achieves the best results overall (\textbf{ARI} = 0.254 and \textbf{{F}$_\beta$} = 0.296 in average), followed by SawETM (\textbf{ARI} = 0.226 and \textbf{{F}$_\beta$} = 0.267 in average). Although SawETM outperforms BoxTM on the arXiv dataset, it cannot discover coherent topics according to the intrinsic evaluation. These results show that there is a tradeoff between learning high-quality topics and document-level semantics for topic modeling methods, and our BoxTM strikes a good balance.

\begin{table}[!t]
\caption{Intrinsic and extrinsic metric scores of ablation models on NYT.}
\label{tab:ablation}
\renewcommand{\arraystretch}{1.15}
\centering
\resizebox{1.0\linewidth}{!}{
\begin{tabular}{|c|c|cccc|}
\hline
\textbf{embedding} & \textbf{model}  & \textbf{C*D} & \textbf{HC} & \textbf{ARI} & \textbf{F$_\beta$} \\
\hline \hline
\multirow{4}{*}{\textbf{box}} &  \textbf{BoxTM} & 0.265 &	0.177 &	\textbf{0.541} &	\textbf{0.577}  \\
\cline{2-6}
& wo/ $\mathcal{L}_{CO}$ & 0.266 &	\textbf{0.191} &	0.449 &	0.489  \\ 
& wo/ $\mathcal{L}_{HT}$ & \textbf{0.276} &	0.157 &	0.299 &	0.355 \\
& wo/ clus & 0.256 &	0.139 &	0.337 &	0.394  \\
\hline
\multirow{4}{*}{\textbf{point}} & w/ kmeans & 0.201 &	0.174 &	0.397 &	0.441  \\
& w/ AP & 0.241 &	0.158 &	0.444 &	0.488  \\
& w/ hier & 0.208 &	0.162 &	0.417 &	0.458  \\ 
& wo/ clus & 0.193 & 0.153 & 0.376 & 0.423 \\
\hline
\end{tabular}}
\end{table}

\subsection{Human Evaluation}
To complement the above automatic metrics, we also utilize a manual evaluation task of \textit{topic intrusion} \cite{chang2009reading} to further validate the ability of topics at different levels to describe documents. As shown in Figure \ref{fig:topic-intrusion} (left), human raters are shown a document from the testing set of NYT, along with four topics represented by their top-10 keywords. Three of them are the top-3 topics at the same level assigned to the given document by the topic model, while the remaining \textit{intruder topic} is sampled randomly from the other low probability topics. We recruit ten graduate students majoring in computer science as raters and instruct them to choose topics that are not relevant to the documents. For evaluation, we compare our BoxTM with two strong baselines, i.e., SawETM and HyperMiner, excluding TaxoGen that cannot infer the topic distributions of documents. According to the value of Light's kappa \cite{light1971kappa} ($\kappa$ = 0.607), the annotation results of the ten raters have a fairly high degree of agreement.

Figure \ref{fig:topic-intrusion} (right) shows the precision scores of different models on this task. The performance of all three models on the manual assessment is generally consistent with those on the extrinsic evaluation. Notably, our BoxTM achieves an overall optimal result, which indicates that it generates different levels of topics that describe documents in alignment with human judgement.

\subsection{Ablation Analysis}
In this section, we conduct an ablation study to analyze the roles of several key components of BoxTM, whose results are shown in Table \ref{tab:ablation}. Most importantly, the ablation models that replace box embeddings with traditional point embeddings (i.e., the \textbf{point} models), experience a drastic performance drop in both topic quality and extrinsic evaluation compared to BoxTM. Within several clustering algorithms, the \textbf{point} model using AP clustering (w/ AP) performs better than those with kmeans++ (w/ kmeans) or agglomerative clustering (w/ hier). 

In terms of the proposed box embedding regularizations, BoxTM wo/ $\mathcal{L}_{HT}$ fails to capture the proper semantic scopes of topics at different levels, leading to worse performance on the \textbf{HC} metric as well as the downstream task. Though BoxTM wo/ $\mathcal{L}_{CO}$ remains competitive on intrinsic evaluation, its performance on the hierarchical text clustering task drops compared to BoxTM.

\begin{figure*}[!t]
\centering
\includegraphics[width=5.5in]{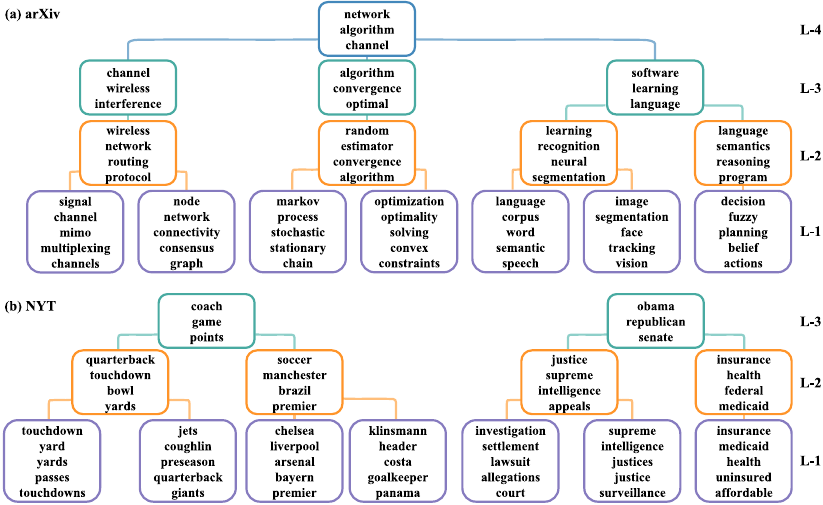}
\caption{Illustration of the partial topic taxonomy learned by BoxTM on arXiv \textbf{(a)} and NYT \textbf{(b)}.}
\label{fig:sample_taxonomy}
\end{figure*}

\subsection{Case Study of Topic Taxonomy}

In this section, we evaluate the mined topic taxonomy qualitatively via a case study. Figure \ref{fig:sample_taxonomy} (a) illustrates some sample topics from the 5-level topic taxonomy learned by BoxTM on the arXiv dataset. A level-4 topic about ``\textit{network}'' branches into child topics related to ``\textit{computer communication networks}'' (left), ``\textit{optimization algorithms}'' (middle), and ``\textit{applications}'' (right). Furthermore, in the field of ``\textit{applications}'', there are sub-fields that focus on different research problems, including ``\textit{computation and language}'' and ``\textit{computer vision and pattern recognition}''. Moreover, Figure \ref{fig:sample_taxonomy} (b) shows some topics related to ``\textit{sports}'' and ``\textit{administration}'' mined by BoxTM on NYT.

\begin{figure}[!t]
\centering
\includegraphics[width=\linewidth]{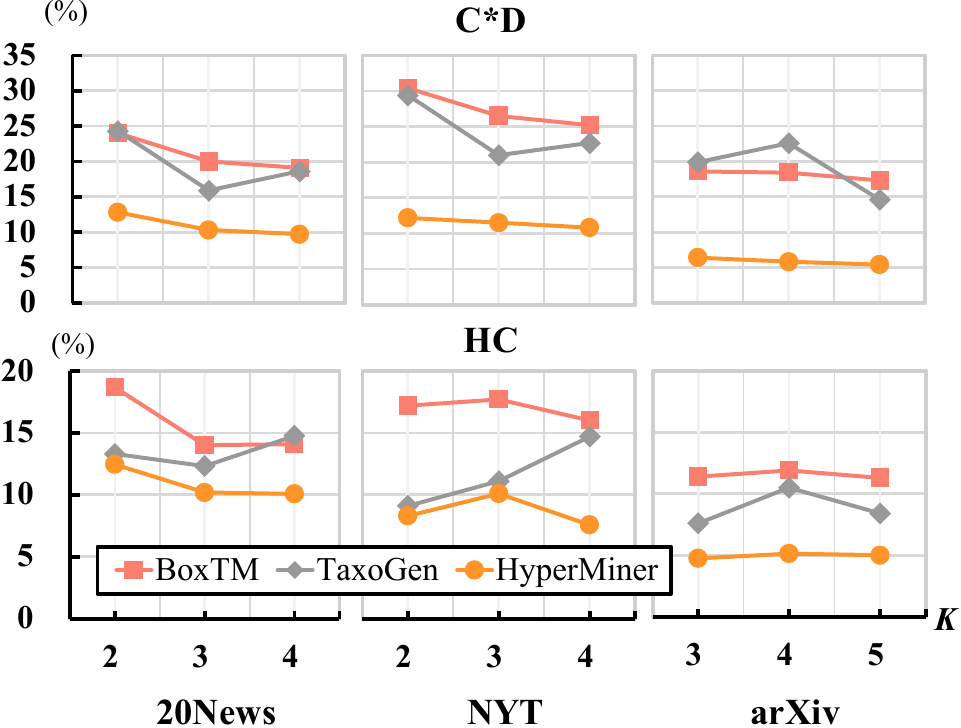}
\caption{The \textbf{C*D} and \textbf{HC} scores of BoxTM, TaxoGen, and HyperMiner with different settings of taxonomy depth (i.e., $K$). }
\label{fig:structure-auto-matric}
\end{figure}

\begin{figure}[!t]
\centering
\includegraphics[width=\linewidth]{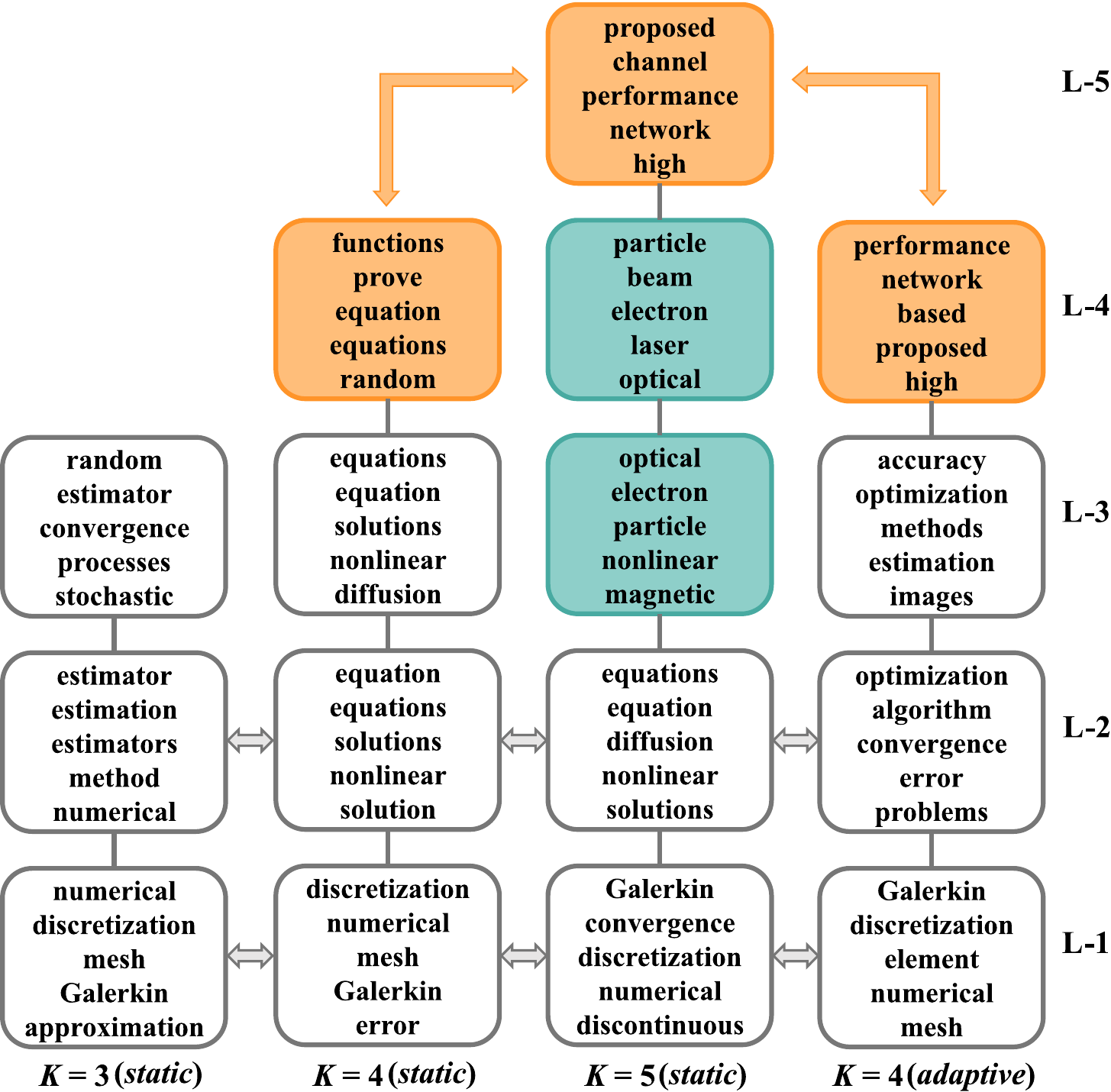}
\caption{Pathways of the leaf topic about ``\textit{Galerkin methods}'' obtained by BoxTM on the arXiv dataset, when the taxonomy depth (i.e., $K$) is set to different values. }
\label{fig:structure-ax}
\end{figure}

\subsection{Analysis of Taxonomy Depth}
In the aforementioned experiments, we set the maximum depth to the same value for all models by following \citet{chen2021tree}. As a complement, Figure \ref{fig:structure-auto-matric} illustrates the performance of our BoxTM compared to the top-2 best performing baselines (i.e., TaxoGen and HyperMiner) for different settings of taxonomy depth. In most cases, BoxTM outperforms baselines with the same taxonomy depth. Nevertheless, how to determine an appropriate taxonomy depth in the real-life applications is a valuable but challenging problem.

Considering that the automatic metrics (e.g., \textbf{C} and \textbf{HC}) may be sensitive to the taxonomy depth, we also conduct a qualitative analysis to discuss the influence of taxonomy depths on our BoxTM. As shown in Figure \ref{fig:structure-ax}, the leaf topic about ``\textit{Galerkin methods}'' is assigned to the parent topic related to ``\textit{numerical analysis}'' for $K$ = 3. And when $K$ = 4, BoxTM further extracts a level-4 topic that is related to ``\textit{general algorithm}''. Interestingly, when the structure of the taxonomy continues to deepen ($K$ = 5), BoxTM identifies that ``\textit{Galerkin methods}'' is commonly applied in the field of ``\textit{physics}'' as a classic PDE solver. Overall, our BoxTM can discover topics with different granularity and the hierarchical relations under varying settings of taxonomy depth. Therefore, users can set the taxonomy depth according to their practical requirements.

Moreover, unlike most HTMs that require a fixed taxonomy depth, the recursive topic clustering module in BoxTM provides a promising solution for determining the taxonomy depth adaptively. Specifically, BoxTM can halt topic clustering when the number of topics at the top level is smaller than a threshold, which is easier to determine compared to the taxonomy depth. Figure \ref{fig:structure-ax} (\textit{adaptive}) illustrates the topic pathway mined by BoxTM when the threshold is set to 10.

\subsection{Qualitative Analysis of Box Embeddings}
In this section, we examine whether box embeddings can reflect the asymmetric relation between parent and child topics. For example, topic 2-5 (i.e., the 5-th topic  at level-2) learned by BoxTM on NYT is related to ``\textit{religion}'' and topic 1-13 is one of its children, while topic 1-27 is about ``\textit{hardware}'', characterized by keywords such as ``\textit{drive}'' and ``\textit{controller}''. As shown in Figure \ref{fig:qualititive} (a), the boxes of upper-level topics entail those of their children. Besides, Figure \ref{fig:qualititive} (b) illustrates that the box embedding of child topic 1-13 has a larger overlap with its parent topic 2-5 compared to a randomly sampled topic 2-11, with $p=0.007<0.05$ according to the paired sample t-test.

\begin{figure}[!t]
\centering
\begin{subfigure}{0.45\linewidth}{
\includegraphics[width=1.4in]{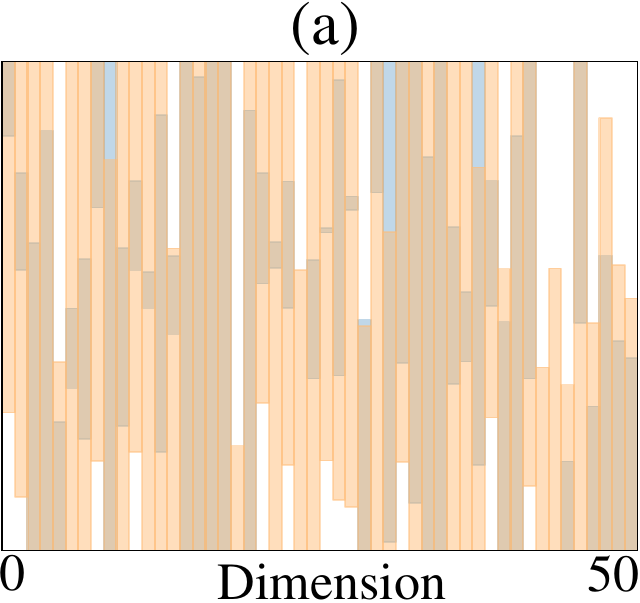}}
\end{subfigure}
\hfil
\begin{subfigure}{0.45\linewidth}{
\includegraphics[width=1.4in]{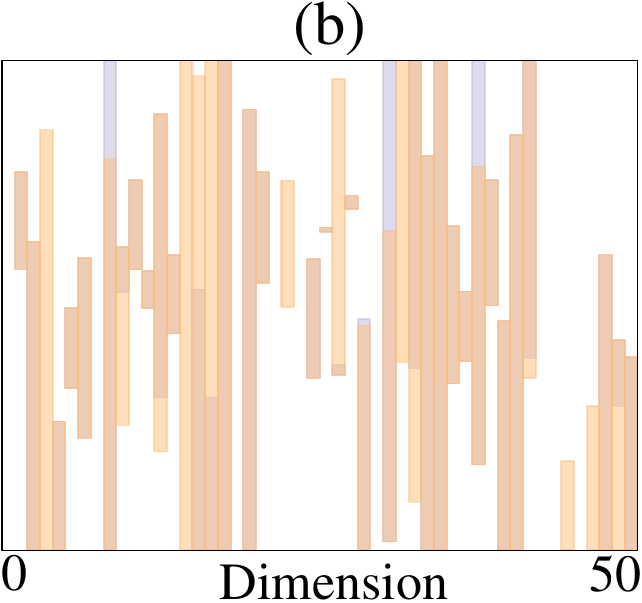}}
\end{subfigure}
\caption{ \textbf{(a)} Visualization of parent topic 2-5 (yellow) and child topic 1-13 (blue) boxes. \textbf{(b)}  Visualization of intersection boxes of hierarchical topics (i.e., 1-13 and 2-5) (yellow) as well as irrelevant topics (i.e., 1-13 and 2-11) (purple).}
\label{fig:qualititive}
\end{figure}

\section{Conclusion}
This paper proposes a novel model named BoxTM for self-supervised topic taxonomy discovery in the box embedding space. Specifically, BoxTM embeds both topics and words into the same box embedding space, where the symmetric and asymmetric metrics are defined to infer the complex relations among topics and words properly. Additionally, instead of initializing topic embeddings randomly, BoxTM uncovers upper-level topics via recursive clustering on topic boxes. 

While our BoxTM has achieved state-of-the-art performance in multiple evaluation experiments, it also exhibits a limitation in efficiency. The \textbf{point} model, a variant of BoxTM that replaces the box embeddings with point embeddings, is trained for 0.22 GPU (GTX 1080 Ti) hour on the {20news} dataset. Due to the extra computation of box operations compared to dot product, BoxTM costs about 1.0 hour, which reveals the research space for efficient computation of box embeddings.

\iftaclpubformat
\else
\subsection{Self-citations}
\label{sec:self-cite}

Citing one's own relevant prior work should be done,  but use the third
person instead of the first person, to preserve anonymity:
\begin{tabular}{l}
Correct: \ex{Zhang (2000) showed ...} \\
Correct: \ex{It has been shown (Zhang, 2000)...} \\
Incorrect: \ex{We (Zhang, 2000) showed ...} \\
Incorrect: \ex{We (Anonymous, 2000) showed ...}
\end{tabular}
\fi

\iftaclpubformat

\section*{Acknowledgments}
We express our profound gratitude to the action editor and reviewers for their valuable comments and suggestions. This research has been supported by the National Natural Science Foundation of China (62372483), the Faculty Research Grants (DB24A4 and DB24C5) of Lingnan University, Hong Kong, the Research Grants Council of the Hong Kong Special Administrative Region, China (UGC/FDS16/E01/19), and the Hong Kong Research Grants Council under the General Research Fund (project no. PolyU 15200021).
\else
\fi

\bibliography{tacl2021}
\bibliographystyle{acl_natbib}

\iftaclpubformat

\onecolumn

\fi

\end{document}